\documentclass[runningheads]{llncs}

\usepackage{graphicx}
\usepackage{booktabs}
\usepackage{array}
\usepackage{amsmath}
\usepackage{amssymb}
\usepackage{url}
\usepackage[hidelinks]{hyperref}

\newcommand{\W}{\textsc{w}}
\newcommand{\G}{\textsc{g}}
\newcommand{\B}{\textsc{b}}

\begin{document}

\title{Embedding-Space Fingerprinting of LoRA Fine-Tunes:\\
Recovering the Training-Data Domain from the Model Artifact Alone}

\titlerunning{Fingerprinting the Training-Data Domain of LoRA Fine-Tunes}

\author{Shivesh Rane\inst{1}}
\authorrunning{S. Rane}

\institute{Goa College of Engineering, Farmagudi, Ponda, Goa, India\\
\email{shiveshrane04@gmail.com}}

\maketitle

\begin{abstract}
When a base language model is fine-tuned with Low-Rank Adaptation (LoRA), how much
of the \emph{training-data domain} leaks into the resulting artifact? This paper
studies whether a simple classifier can recover the domain of a fine-tune given
\emph{only} the model artifact (weight deltas, internal activations on a fixed probe
set, or generated text) without ever observing the training data. Using one base
model (Qwen3.5-0.8B), 36 LoRA fine-tunes are produced over 18 public datasets in five
domains (medical, code, mathematics, non-English text, narrative prose), and three
feature families are extracted at decreasing levels of access: white-box weights,
gray-box activations, and black-box outputs. A leave-one-dataset-out (LODO) protocol,
frozen before any result was seen, tests whether the domain of a fine-tune trained on
a \emph{previously unseen dataset} can be identified. The domain is recoverable well
above chance (20\%) from gray-box activations (61\%) and black-box outputs (64\%);
white-box weight statistics reach 42\%. Contrary to the pre-registered hypothesis that
signal decays monotonically from weights to activations to outputs, activations and
outputs were \emph{more} informative than the weight-delta statistics. The black-box
result holds unchanged (64\%) when only generated text is used, without token
probabilities. Domain accuracy is also stable when a third run per dataset is trained
on an independently drawn data subsample, and black-box predictions are unaffected by
the LoRA rank (8--64). All code and per-run feature vectors are released.
\end{abstract}

\section{Introduction}
\label{sec:intro}
Parameter-efficient fine-tuning, and LoRA~\cite{hu2022lora} in particular, has made
it cheap to specialize a shared base model to a particular data distribution and to
distribute the resulting adapter as a small artifact; quantized variants such as
QLoRA~\cite{dettmers2023qlora} bring this within reach of a single commodity GPU, and
adapter libraries are routinely shared and composed~\cite{huang2024lorahub}. This
raises a provenance question with both auditing and privacy dimensions: \emph{given
only the artifact}, can one infer what \emph{kind} of data the adapter was trained on?

The study is organized around three research questions, in decreasing order of access
to the model:
\begin{description}
  \item[RQ1 (per-dataset).] Can a classifier identify \emph{which} dataset produced a
  fine-tune? (a re-identification baseline)
  \item[RQ2 (per-domain, the headline).] Can it identify the \emph{domain} of a
  fine-tune trained on a dataset from which it has seen \emph{no} examples, i.e.,
  does the fingerprint \emph{generalize} across datasets within a domain?
  \item[RQ3 (signal decay).] How does recoverability change as access degrades from
  weight deltas, to activations on a fixed probe set, to generated outputs?
\end{description}

The contributions are:
(i) a formulation of training-domain inference for LoRA adapters along an explicit
white/gray/black-box access axis (Sect.~\ref{sec:formulation});
(ii) an end-to-end empirical study on a modern hybrid base model under an evaluation
protocol frozen before any fine-tune was trained (Sects.~\ref{sec:method}
and~\ref{sec:results});
(iii) a clearly characterized outcome, including a refuted sub-hypothesis (RQ3), with
dataset-clustered confidence intervals; and
(iv) robustness checks on the black-box access assumption
(Sect.~\ref{sec:robust-textonly}), the LoRA rank (Sect.~\ref{sec:robust-rank}), and
independently resampled training data (Sect.~\ref{sec:robust-resample}).

\section{Related Work}
\label{sec:related}

\paragraph{Learning from weights and fine-tuned representations.}
Model properties can be predicted directly from parameters, e.g., test accuracy from
weights~\cite{unterthiner2020predicting}, and tasks can be embedded as vectors, e.g.,
Task2Vec~\cite{achille2019task2vec}, an ancestor of the white-box features used here.
Recent work treats LoRA factors as a data modality: Putterman et
al.~\cite{putterman2024lol} build GL-equivariant networks that predict attributes and
membership of the fine-tuning data from LoRA weights, Duszenko and
Bielak~\cite{duszenko2025weightspace} analyze how much concept information lives in
diffusion-model adapter weights, and Horwitz et al.\ recover pre-fine-tuning weights
from several LoRA fine-tunes~\cite{horwitz2024spectral} and the heritage tree of
fine-tuned models from their weights~\cite{horwitz2025mother}. Closest to this paper,
Delta Activations~\cite{xu2025delta} embeds LoRA fine-tunes of 7--9B models by the
shift of their final-layer activations relative to the base model on generic prompts,
and shows that the embeddings \emph{cluster} by domain, while flattened weights and
output sentence embeddings cluster poorly. The present study differs in three ways:
it asks a \emph{supervised generalization} question (predicting the domain of a
fine-tune from a dataset never seen by the classifier), it compares the three access
levels under one frozen protocol, and it finds output features competitive with
activations rather than inferior to them.

\paragraph{Membership, dataset, and property inference.}
Membership inference~\cite{shokri2017membership} asks whether a specific example was
used for training; for large language models (LLMs) it is often barely above
chance~\cite{duan2024mia}, and token-probability heuristics such as Min-K\%
Prob~\cite{shi2024minkprob} target pretraining-data detection. Dataset
inference~\cite{maini2021dataset} aggregates weak per-example signals to decide
whether a \emph{given} dataset was used, and has been extended to
LLMs~\cite{maini2024llmdi}. Property inference on fine-tuned LLMs recovers
dataset-level statistics such as patient demographics~\cite{huang2025propinfer}. All
of these assume a candidate dataset or a known property; the present question is
coarser and requires no candidate data: \emph{which domain} the adapter was trained on.

\paragraph{Model fingerprinting.}
Fingerprinting usually identifies \emph{which model} an artifact derives from, either
by implanting a secret fingerprint~\cite{xu2024instructional} or by exploiting
\emph{natural} fingerprints left by training dynamics, which make models trained on
identical data distinguishable~\cite{suzuki2025natural}. The latter is directly
relevant to RQ1, where same-dataset runs are re-identified almost perfectly
(Sect.~\ref{sec:results}). Here the target of the fingerprint is the
\emph{training data}, not the model lineage.

\section{Problem Formulation}
\label{sec:formulation}
Let $M_{\mathrm{base}}$ be a fixed base model and $\mathcal{D}=\{D_1,\dots,D_K\}$ a set
of datasets, each mapped to one of $C$ domains by a labeling
$c:\mathcal{D}\to\mathcal{Y}$, $|\mathcal{Y}|=C$ (here $C=5$). For dataset $D_i$ and
random seed $s$ a fine-tune is obtained as
\begin{equation}
  M_{i,s} = \mathrm{FT}\left(M_{\mathrm{base}},\,D_i;\,s\right),
  \label{eq:ft}
\end{equation}
under one fixed training recipe (Sect.~\ref{sec:population}). For an access level
$a\in\{\W,\G,\B\}$ a feature map $\phi_a$ turns the \emph{artifact alone} into a
fixed-length vector
\begin{equation}
  \mathbf{x}^{(a)}_{i,s} = \phi_a\left(M_{i,s}\right)\in\mathbb{R}^{d_a},
  \label{eq:featmap}
\end{equation}
using no access to $D_i$. The three access levels are defined by what $\phi_a$ may
query:
\begin{itemize}
  \item \textbf{White-box (\W):} the adapter parameters (the LoRA factors).
  \item \textbf{Gray-box (\G):} hidden states of every layer of $M_{i,s}$ on chosen
  inputs, but not its parameters.
  \item \textbf{Black-box (\B):} the generated text for chosen inputs \emph{together
  with the per-step next-token probability distribution} (as exposed by APIs that
  return log-probabilities). A strictly \emph{text-only} variant, which drops every
  feature that needs token probabilities, is evaluated separately
  (Sect.~\ref{sec:robust-textonly}).
\end{itemize}
The goal is a classifier $g_a:\mathbb{R}^{d_a}\to\mathcal{Y}$ that predicts the domain
$c(D_i)$. The headline question (RQ2) requires \emph{generalization across datasets}:
$g_a$ is evaluated on fine-tunes of datasets whose examples never appear in its
training set (Sect.~\ref{sec:eval}).

\section{Method}
\label{sec:method}

\subsection{Pipeline Overview}
The study proceeds in four stages, each fully scripted and released:
\begin{enumerate}
  \item \emph{Freeze the evaluation}: commit the dataset-to-domain map, the 200 probe
  prompts, the LODO folds, and the scoring script before any training.
  \item \emph{Train}: fine-tune $M_{\mathrm{base}}$ once per (dataset, seed) with an
  identical LoRA recipe (Eq.~\eqref{eq:ft}), saving only the adapter.
  \item \emph{Extract}: compute the feature vectors of Eq.~\eqref{eq:featmap}
  for every fine-tune, as described in Sect.~\ref{sec:features}.
  \item \emph{Score}: run the frozen script once (Sect.~\ref{sec:eval}).
\end{enumerate}

\subsection{Population of Fine-Tunes}
\label{sec:population}
All fine-tunes share one base model, \texttt{Qwen3.5-0.8B} (a hybrid
linear-attention vision-language model, used text-only), loaded in 4-bit NF4
quantization~\cite{dettmers2023qlora}. Table~\ref{tab:datasets} lists the five domains
and their datasets. Every run uses the same recipe: LoRA with rank $r=16$ and
$\alpha=32$ on all linear projections inside the token-mixing and MLP blocks
(introspected rather than hard-coded, so that the gated-delta projections of the hybrid
architecture are included); a fixed budget of $1.5$M training tokens, so that dataset
size is not a confound (one exception: the pipeline-validation run, PubMedQA seed 0,
was trained on $2.0$M tokens before the budget was fixed and was kept; see
Sect.~\ref{sec:limits}); maximum sequence length 1024 with packing; learning rate
$2\times10^{-4}$ with linear decay; effective batch size 16; one epoch; fp32 compute;
two random seeds per dataset. Everything except the training dataset is held constant.
Two of the twenty intended datasets were excluded before any of their features reached
a classifier (Sect.~\ref{sec:limits}), leaving 18 datasets $\times$ 2 seeds $=$ 36
fine-tunes. Within a dataset, both seeds draw the same token subsample and differ only
in initialization and data order; Sect.~\ref{sec:robust-resample} relaxes this.

\begin{table}[t]
\centering
\caption{Datasets per domain (Hugging Face identifiers abbreviated). Entries in parentheses
were excluded before any of their features reached a classifier
(Sect.~\ref{sec:limits}).}
\label{tab:datasets}
\begin{tabular}{@{}l>{\raggedright\arraybackslash}p{0.74\textwidth}@{}}
\toprule
Domain & Datasets \\
\midrule
Medical      & PubMedQA, MedMCQA, ChatDoctor-HealthCareMagic, MedQA-USMLE \\
Code         & Python-code-instructions-18k, Evol-Instruct-Code-80k, SQL-create-context, Code-instructions-122k \\
Mathematics  & MetaMathQA, MathInstruct, Orca-Math-200k, \textit{(camel-ai/math)} \\
Multilingual & IITB English--Hindi, OPUS-100 de--en, Aya (non-English slice), \textit{(Alpaca-GPT4-Japanese)} \\
Creative     & WritingPrompts, TinyStories, Opus-WritingPrompts, Instruction-Poems \\
\bottomrule
\end{tabular}
\end{table}

\subsection{Feature Families}
\label{sec:features}
The same 200 probe prompts are used for the gray- and black-box families. They are
deliberately domain-neutral (no medical, code, numeric, non-English, or narrative
vocabulary, e.g., ``Describe your morning routine.'') so that any recovered signal
reflects the fine-tune and not leakage from the prompt. Probes are formatted with the
model's chat template.

\paragraph{White-box (weights, $d_{\W}=48$).}
For each adapted module $m$ the LoRA update is
\begin{equation}
  \Delta W_m = \frac{\alpha}{r}\,B_m A_m,\qquad
  B_m\in\mathbb{R}^{o\times r},\; A_m\in\mathbb{R}^{r\times d}.
  \label{eq:deltaw}
\end{equation}
$\Delta W_m$ is never materialized. With thin QR factorizations $B_m=Q_BR_B$ and
$A_m^{\top}=Q_AR_A$,
\begin{equation}
  \Delta W_m = Q_B\left(\tfrac{\alpha}{r}R_BR_A^{\top}\right)Q_A^{\top},
  \label{eq:core}
\end{equation}
and because $Q_B$ and $Q_A$ in Eq.~\eqref{eq:core} have orthonormal columns, the singular values
$\sigma_1\ge\dots\ge\sigma_r$ of $\Delta W_m$ equal those of the $r\times r$ core
$\tfrac{\alpha}{r}R_BR_A^{\top}$. The Frobenius norm follows as
\begin{equation}
  \lVert\Delta W_m\rVert_F = \Big(\textstyle\sum_{k=1}^{r}\sigma_k^2\Big)^{1/2}.
  \label{eq:frob}
\end{equation}
Per module the norm (Eq.~\eqref{eq:frob}) and the top-8 singular values are kept and
aggregated into a fixed-length vector: mean, standard deviation, maximum, and sum of
the norms per block type (token-mixing, MLP, other; 12 values); the normalized layer
profile
\begin{equation}
  p(\ell) = \frac{\sum_{m\in\text{layer }\ell}\lVert\Delta W_m\rVert_F}
                 {\sum_{\ell'}\sum_{m\in\text{layer }\ell'}\lVert\Delta W_m\rVert_F},
  \label{eq:profile}
\end{equation}
resampled to 24 points; the normalized mean top-8 spectrum (8 values); and four global
scalars (log total norm, token-mixing:MLP norm ratio, early:late layer norm ratio, and
module count).

\paragraph{Gray-box (activations, $d_{\G}=78$).}
For probe $p$ with $T_p$ tokens and hidden states $\mathbf{h}_{\ell,t}(p)$ at layer
$\ell$, the per-layer mean and within-prompt spread are averaged over the probe set
$\mathcal{P}$:
\begin{equation}
  \boldsymbol{\mu}_\ell = \frac{1}{|\mathcal{P}|}\sum_{p\in\mathcal{P}}
  \frac{1}{T_p}\sum_{t=1}^{T_p}\mathbf{h}_{\ell,t}(p),\qquad
  \mathbf{s}_\ell = \frac{1}{|\mathcal{P}|}\sum_{p\in\mathcal{P}}
  \mathrm{std}_t\big(\mathbf{h}_{\ell,t}(p)\big).
  \label{eq:acts}
\end{equation}
From Eq.~\eqref{eq:acts}, three layer curves, $\lVert\boldsymbol{\mu}_\ell\rVert_2$, the mean of
$|\boldsymbol{\mu}_\ell|$, and the mean of $\mathbf{s}_\ell$, are each resampled to
24 points (72 values), plus six global scalars (curve means and spreads, layer count,
and the relative depth of the peak activation norm).

\paragraph{Black-box (outputs, $d_{\B}=455$).}
Each probe is answered by greedy decoding (at most 64 new tokens). With $p_t$ the
next-token distribution at generation step $t$, the mean predictive entropy per probe is
\begin{equation}
  \bar{H}(p) = \frac{1}{T}\sum_{t=1}^{T} H_t,\qquad
  H_t = -\sum_{v\in\mathcal{V}} p_t(v)\log p_t(v).
  \label{eq:entropy}
\end{equation}
The feature vector concatenates the mean, standard deviation, and median of
$\bar{H}(p)$ over probes (3 values, the only features requiring token probabilities);
response-length statistics (4); a 64-bucket hashed histogram of the 200 most frequent
generated token ids (64); and the centroid of the Sentence-BERT~\cite{reimers2019sbert}
embeddings (all-MiniLM-L6-v2) of the generations (384).

\subsection{Frozen Evaluation Protocol}
\label{sec:eval}
The evaluation was \emph{frozen before any fine-tune was trained}: the
dataset-to-domain map, the probes, the folds, and the scoring script were committed and
never altered. For RQ2, leave-one-dataset-out (LODO) is used. Writing
$\mathcal{R}=\{(i,s)\}$ for all runs and $g_a^{(-i)}$ for the classifier trained on
every run \emph{except} those of dataset $D_i$, the pooled LODO accuracy is
\begin{equation}
  \mathrm{Acc}_a = \frac{1}{|\mathcal{R}|}\sum_{(i,s)\in\mathcal{R}}
  \mathbf{1}\left[\,g_a^{(-i)}\big(\mathbf{x}^{(a)}_{i,s}\big)=c(D_i)\,\right],
  \label{eq:lodo}
\end{equation}
so the domain of $D_i$ must be inferred purely from \emph{other} datasets of that
domain. Each $g_a$ is a standardized logistic regression (inverse regularization
strength 1). RQ1 instead uses leave-one-seed-out and predicts the dataset identity.
Reported metrics are top-1 accuracy, macro-F1, and a 1000-resample bootstrap 95\%
percentile confidence interval (CI) over pooled predictions. Because the two seeds of a
dataset share training data, their predictions are not independent; a
\emph{dataset-clustered} bootstrap, which resamples whole datasets, is therefore also
reported. Chance for the five-way domain task is $1/C=0.20$. Feature development used
only a disjoint seed-0 development split, and the frozen script was executed once on
the 36-run population.

\section{Results}
\label{sec:results}
\begin{figure}[t]
\centering
\includegraphics[width=0.62\textwidth]{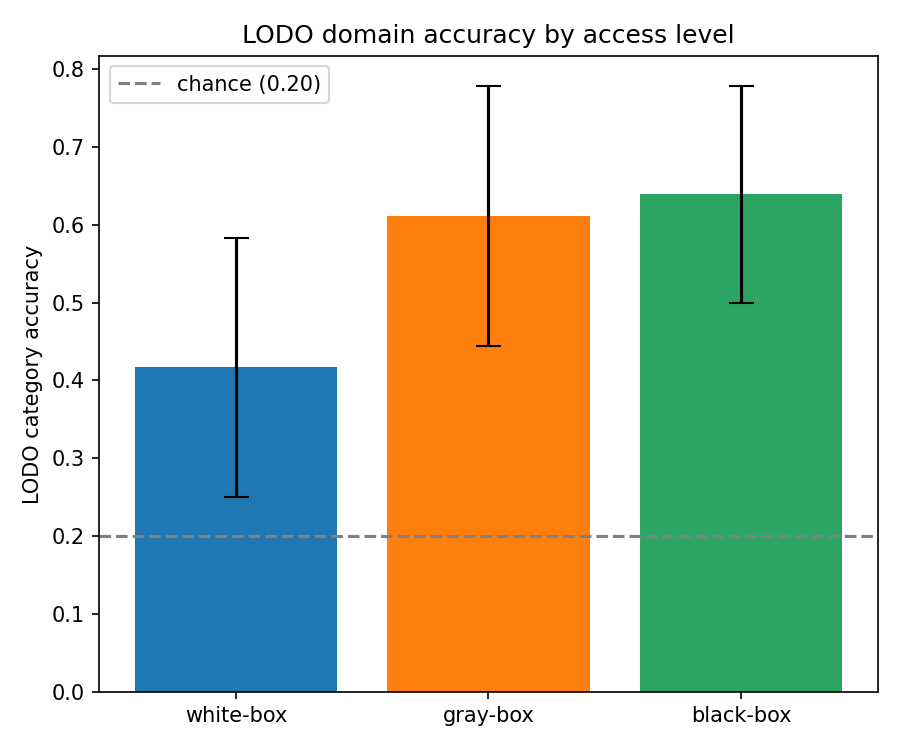}
\caption{RQ2 leave-one-dataset-out domain accuracy by access level, with bootstrap
95\% CIs. The dashed line is chance (0.20).}
\label{fig:bars}
\end{figure}

Table~\ref{tab:main} gives the headline numbers, with the RQ2 accuracy computed
as in Eq.~\eqref{eq:lodo}; Fig.~\ref{fig:bars} plots these accuracies against chance.

\begin{table}[t]
\centering
\caption{Frozen LODO results (36 runs). RQ2 predicts the domain of a fine-tune from
datasets never seen in training (chance $=0.20$); the row CI resamples individual
predictions and the clustered CI resamples whole datasets. RQ1 re-identifies the
dataset across seeds (chance $=0.056$).}
\label{tab:main}
\begin{tabular}{lccccc}
\toprule
 & \multicolumn{4}{c}{RQ2: domain of an unseen dataset} & RQ1 \\
\cmidrule(lr){2-5}
Access level & Acc. & 95\% CI (row) & 95\% CI (clustered) & Macro-F1 & Acc. \\
\midrule
White-box (weights)    & 0.417          & 0.250--0.583 & 0.194--0.639 & 0.404 & 1.000 \\
Gray-box (activations) & \textbf{0.611} & 0.444--0.778 & 0.417--0.806 & 0.643 & 1.000 \\
Black-box (outputs)    & \textbf{0.639} & 0.499--0.778 & 0.417--0.861 & 0.640 & 0.944 \\
\bottomrule
\end{tabular}
\end{table}

\paragraph{RQ1 (per-dataset).}
Re-identifying the exact dataset across seeds is essentially perfect from weights and
activations (1.000) and nearly so from outputs (0.944, chance 0.056)
(Table~\ref{tab:main}, last column). Since both seeds of a dataset share their training
subsample, this mainly shows that the artifact encodes its specific training set, in
line with natural-fingerprint findings~\cite{suzuki2025natural}; it is not evidence of
domain learning, which is what RQ2 measures.

\paragraph{RQ2 (per-domain, the headline).}
The domain of a fine-tune is recoverable on \emph{unseen} datasets. Gray-box
activations reach 0.611 and black-box outputs 0.639, with both row and clustered CIs
excluding chance (Table~\ref{tab:main}, Fig.~\ref{fig:bars}). Weight statistics reach
0.417; the row CI excludes chance but the more conservative clustered CI
(0.194--0.639) does not, so the white-box result alone should not be read as
established.

\begin{figure}[t]
\centering
\includegraphics[width=0.6\textwidth]{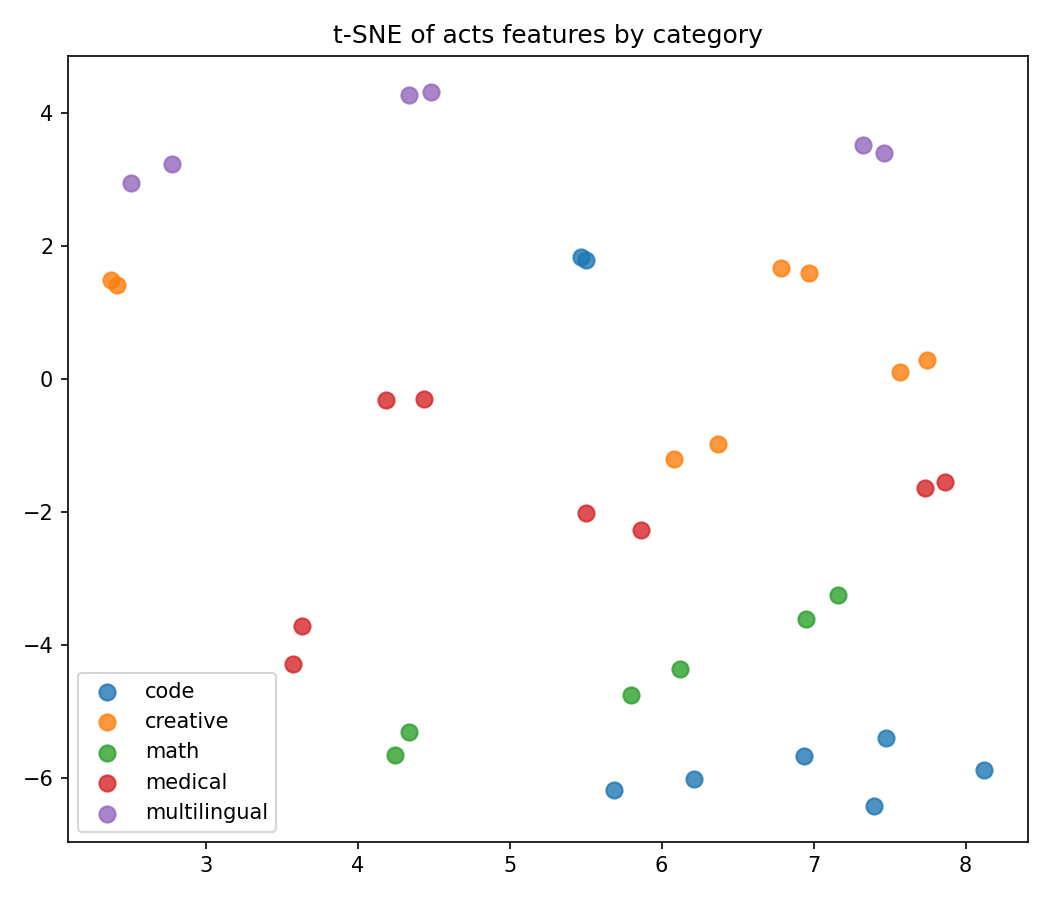}
\caption{t-SNE of the gray-box activation features, one point per fine-tune, colored
by training-data domain.}
\label{fig:tsne}
\end{figure}

Figure~\ref{fig:tsne} visualizes the gray-box features with t-SNE. The two seeds of
every dataset appear as a tight pair, the geometric counterpart of the perfect RQ1
accuracy. At the domain level the layout is only partly organized: multilingual
fine-tunes occupy the top of the map and mathematics and most code fine-tunes the
bottom, while medical and creative fine-tunes are interleaved in the middle. This
partial structure is consistent with an RQ2 accuracy well above chance but far from
perfect, and anticipates the main confusion below.

\begin{figure}[t]
\centering
\includegraphics[width=0.48\textwidth]{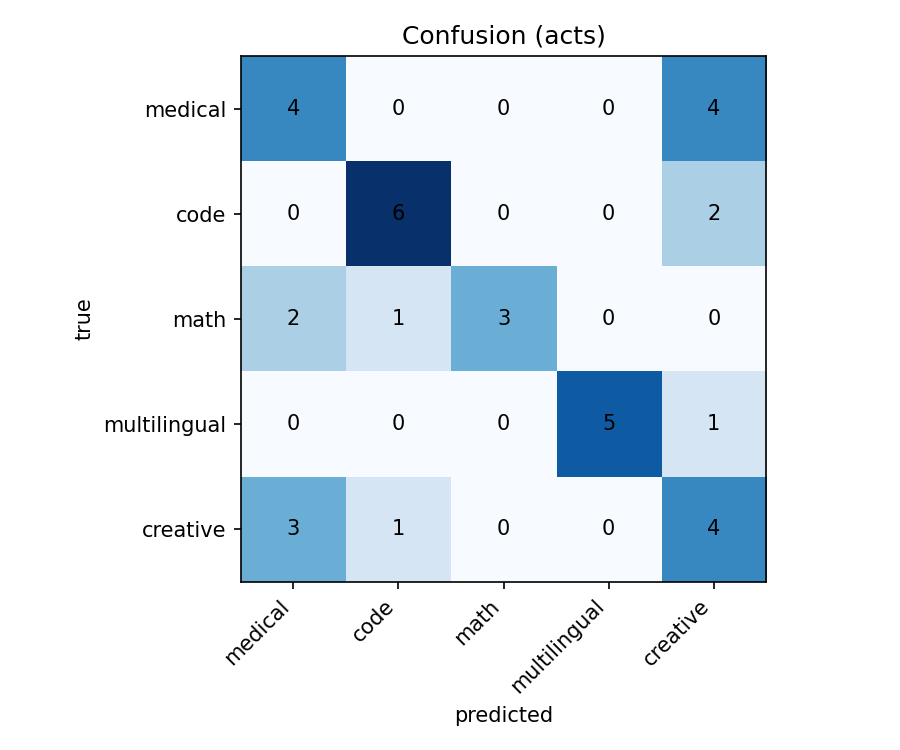}\hfill
\includegraphics[width=0.48\textwidth]{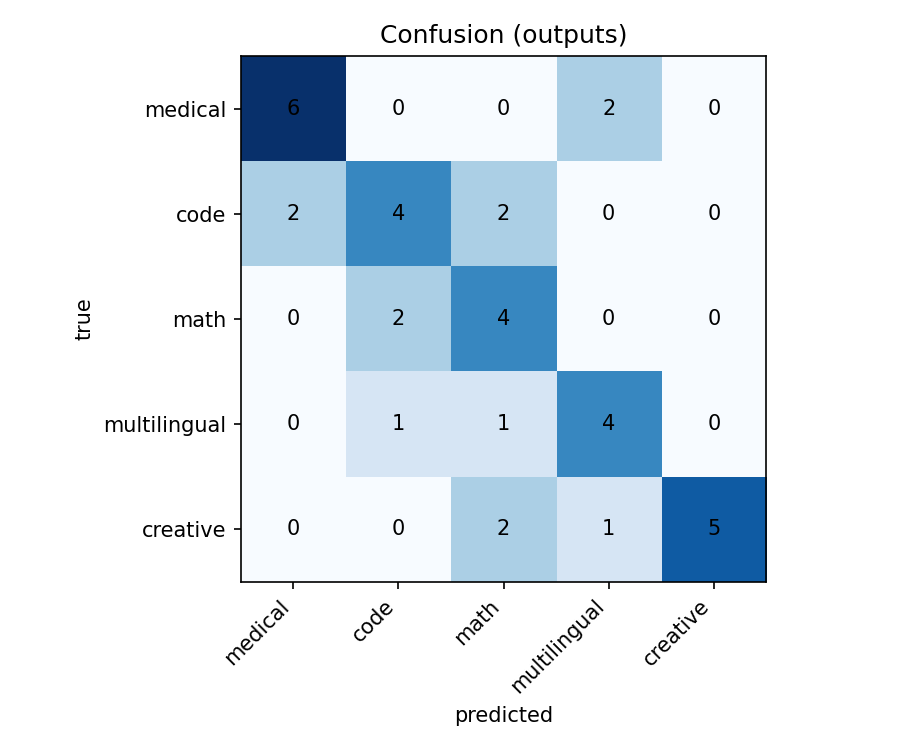}
\caption{Pooled LODO confusion matrices for gray-box activations (left) and black-box
outputs (right). Rows are true domains, columns predicted domains.}
\label{fig:conf}
\end{figure}

The pooled confusion matrices in Fig.~\ref{fig:conf} show where the errors occur. With
activations (left), multilingual (5 of 6 runs correct) and code (6 of 8) are the most
separable domains, while medical and creative fine-tunes are confused with each other
(4 of 8 medical runs predicted creative and 3 of 8 creative runs predicted medical):
both domains consist of English prose, which shifts internal statistics similarly. With
outputs (right), the medical--creative confusion disappears (6 of 8 medical and 5 of 8
creative runs correct); the largest remaining confusion is between code and mathematics
(4 of the 13 errors), two domains whose training data are both dominated by symbolic,
step-by-step answers. The two
access levels therefore fail on different domain pairs, which suggests they capture
complementary aspects of the fine-tune.

\paragraph{RQ3 (signal decay).}
The pre-registered hypothesis was that recoverability decays monotonically,
white~$>$~gray~$>$~black. The data \emph{refute} this ordering: activations (0.611) and
outputs (0.639) were more informative than the weight-delta statistics (0.417). Either
the domain imprints more on behavior than on delta magnitudes, or, more likely, the
hand-designed weight features of Sect.~\ref{sec:features} (norms, spectra, and layer
profiles, Eqs.~\eqref{eq:frob}--\eqref{eq:profile}, all of which discard singular
\emph{directions}) leave recoverable structure unused. This agrees
with the weak clustering of raw adapter weights reported for larger
models~\cite{xu2025delta}, but differs from that study on outputs: there, output
sentence embeddings clustered poorly, whereas here output features are the strongest
family. The difference does not come from the feature set: the MiniLM sentence-embedding
centroid \emph{alone}, the same output representation that study used, reaches 0.611
LODO accuracy here (Sect.~\ref{sec:robust-textonly}). Setting aside the other
differences between the two studies (base models, probe sets), the most likely
explanation is the evaluation: domain information is present in output embeddings but is not the dominant
direction of variation, so a supervised classifier on standardized features recovers it
while unsupervised clustering does not.

\section{Robustness Extensions}
\label{sec:robust}
The checks below were added after acceptance in response to reviewer comments. Each is
scored by the same unmodified frozen script on a separate set of runs, so the
pre-registered results of Table~\ref{tab:main} are unaffected. Table~\ref{tab:robust}
summarizes them. For these checks all runs were re-extracted with a deterministic hash
for the token-histogram features (the released code); on the original 36 runs this
reproduces the published accuracies of Table~\ref{tab:main} exactly.

\begin{table}[t]
\centering
\caption{Post-acceptance robustness checks (RQ2 LODO domain accuracy unless noted;
chance $=0.20$).}
\label{tab:robust}
\begin{tabular}{lccc}
\toprule
Setting & White-box & Gray-box & Black-box \\
\midrule
Headline (Table~\ref{tab:main}), 36 runs        & 0.417 & 0.611 & 0.639 \\
Text-only black-box (no entropy), 36 runs        & --    & --    & 0.639 \\
MiniLM centroid only (384 dims), 36 runs         & --    & --    & 0.611 \\
Rank transfer, $r\in\{8,64\}$ (correct / 10)     & 3/10  & 6/10  & 8/10  \\
Three runs per dataset incl.\ resampled, 54 runs & 0.407 & 0.648 & 0.630 \\
RQ1 across independent subsamples, 54 runs       & 1.000 & 0.981 & 0.981 \\
\bottomrule
\end{tabular}
\end{table}

\subsection{Text-Only Black-Box Access}
\label{sec:robust-textonly}
The black-box family of Sect.~\ref{sec:features} needs token probabilities for the
three entropy features (Eq.~\eqref{eq:entropy}). Removing them leaves a 452-dimensional
vector computable from generated text alone. On the same 36 runs its LODO accuracy is
0.639 (row CI 0.499--0.778, clustered CI 0.417--0.861; macro-F1 0.641), identical to
the full black-box family, and RQ1 is unchanged at 0.944 (Table~\ref{tab:robust});
only one of the 36 predictions differs. The black-box signal therefore does not depend
on token-probability access: generated text alone suffices.
Going further, the 384-dimensional MiniLM centroid alone, without length or token
statistics, still reaches 0.611 (row CI 0.444--0.778, clustered CI 0.389--0.833;
macro-F1 0.620), so most of the black-box signal is carried by the semantic content of
the generations.

\subsection{LoRA Rank}
\label{sec:robust-rank}
Ten additional fine-tunes use $r\in\{8,64\}$ (with $\alpha=2r$, so the scale
$\alpha/r$ of Eq.~\eqref{eq:deltaw} is unchanged) on one dataset per domain. A
classifier trained only on rank-16 runs of the \emph{other} datasets predicts the
domain of each rank-8 and rank-64 run, which tests rank transfer and dataset transfer
together. Correct predictions are 3/10, 6/10, and 8/10 for the white-, gray-, and
black-box families (Table~\ref{tab:robust}); for reference, the same five datasets at
rank 16 give 3/5, 2/5, and 4/5. The black-box family is unaffected by rank: every
dataset receives the same prediction at ranks 8, 16, and 64 (four of five correct,
PubMedQA misclassified at every rank). The white-box statistics are not rank-invariant:
all five rank-64 adapters are predicted as medical (correctly only for PubMedQA),
because spectra and norms shift with the rank itself. When rank-16 runs of the same dataset are allowed in
training (the frozen script's robustness probe), category accuracy on the ten runs is
0.40, 0.80, and 1.00, respectively. With five datasets per rank these counts are
indicative only.

\subsection{Independently Resampled Training Data}
\label{sec:robust-resample}
In the headline population the two seeds of a dataset share one training subsample.
A third run per dataset (18 runs) is therefore trained on an independently drawn
1.5M-token subsample with a new seed. For 15 of the 18 datasets the budget uses at most
26\% of the dataset, so the new subsample is largely disjoint from the original one;
for the other three (Python-code-instructions 46\%, MedQA-USMLE 64\%, Aya 93\%) the
overlap is substantial. On all 54 runs the LODO domain accuracy
is 0.407, 0.648, and 0.630 for the white-, gray-, and black-box families (row CIs
0.278--0.537, 0.519--0.778, 0.519--0.759; clustered CIs 0.203--0.611, 0.426--0.852,
0.407--0.833), essentially unchanged from the 36-run headline but with narrower row
intervals (Table~\ref{tab:robust}). Leave-one-seed-out dataset re-identification, now
across runs trained on largely different examples, stays at 1.000, 0.981, and 0.981
(chance 0.056). The fingerprint therefore characterizes the dataset's distribution
rather than the specific examples drawn from it.

\section{Discussion, Limitations, and Threats to Validity}
\label{sec:limits}
The headline finding, that the training \emph{domain} of a LoRA fine-tune is
recoverable from the artifact alone and generalizes to unseen datasets, bears directly
on provenance auditing of distributed adapters and on detecting undisclosed training
data. The black-box result is notable because, as the text-only variant of
Sect.~\ref{sec:robust-textonly} shows, it requires nothing beyond the generated text
for a fixed set of neutral prompts, i.e., ordinary API access.

The study has clear bounds.
(i) It uses a \emph{single} 0.8B hybrid vision-language base model and LoRA only;
transfer to other architectures, larger scales, and full fine-tuning is untested and
was outside the compute budget of the study (a hard limit of US\$30 of cloud GPU time).
(ii) With 36 headline runs the CIs are wide; under the clustered bootstrap the
white-box accuracy is not distinguishable from chance, so the RQ3 ordering should be
read as suggestive; adding a third run per dataset (Sect.~\ref{sec:robust-resample})
narrows the intervals but does not change this conclusion.
(iii) Two of the twenty intended datasets were excluded before any of their features
reached a classifier: \texttt{camel-ai/math}, which is stored as tens of thousands of
individual files and could not be loaded within the compute budget, and a Japanese
dataset that was deprioritized under the budget; mathematics and multilingual
therefore use three datasets each.
(iv) One of the 36 headline runs (PubMedQA, seed 0) was trained on 2.0M rather than
1.5M tokens, because it served as the pipeline-validation run before the budget was
reduced. Replacing it with a 1.5M-token rerun leaves the RQ2 accuracies unchanged
(0.417, 0.611, 0.639).
(v) The domains are deliberately well separated; finer distinctions (e.g., two
medical sub-specialties) are likely harder.
(vi) A pilot result of 100\% accuracy under leave-one-\emph{seed}-out was shown by a
label-shuffle control to be same-dataset re-identification rather than domain
learning; RQ2 therefore relies solely on leave-one-\emph{dataset}-out, which a held-out
dataset's second seed cannot inflate.

\section{Conclusion}
\label{sec:conclusion}
A simple classifier can recover the training-data \emph{domain} of LoRA fine-tunes of
a single base model from the artifact alone, and the signal transfers to datasets the
classifier never saw, most strongly from activations and generated outputs and more
weakly from weight-delta statistics. The result holds with generated text alone, with
a third run per dataset on independently drawn data, and, for the black-box family,
across LoRA ranks 8 to 64. The result is a starting point for practical
provenance and data-leakage auditing of distributed adapters. Richer white-box features
that use singular directions, and replication on further base models and fine-tuning
methods, are the clear next steps. All code, the frozen evaluation, and per-run feature
vectors are released for reproduction.


\end{document}